\documentclass{article}

\usepackage{arxiv}
\usepackage{amsmath}

\usepackage[utf8]{inputenc} % allow utf-8 input
\usepackage[T1]{fontenc}    % use 8-bit T1 fonts
\usepackage{hyperref}       % hyperlinks
\usepackage{url}            % simple URL typesetting
\usepackage{booktabs}       % professional-quality tables
\usepackage{amsfonts}       % blackboard math symbols
\usepackage{nicefrac}       % compact symbols for 1/2, etc.
\usepackage{microtype}      % micro-typography
\usepackage{graphicx}       % images
\usepackage[dvipsnames]{xcolor}         % colors
\usepackage{enumitem}       % control over itemize and enumerate
\usepackage{newtxtext}      % times font
\usepackage{newtxmath}      % times font (math version)

\newcommand{\datafile}[1]{\texttt{\textbf{#1}}}
\newcommand{\textdatafile}[1]{\texttt{#1}}
\newcommand{\dataitem}[1]{\texttt{#1}}

\newcommand{\textapprox}{\raisebox{0.5ex}{\texttildelow}}

\title{Bonn Activity Maps: Dataset Description}

\newcommand{\authorwidth}{0.3\textwidth}

\author{
Julian Tanke\\
Computer Vision Group\\
University of Bonn\\
\texttt{tanke@iai.uni-bonn.de}\\
\And
Oh-Hun Kwon\\
Computer Vision Group\\
University of Bonn\\
\texttt{s6ohkwon@uni-bonn.de}\\
\And
Patrick Stotko\\
Computer Graphics Group \\
University of Bonn\\
\texttt{stotko@cs.uni-bonn.de}\\
\AND
Radu A. Rosu\\
Autonomous Intelligent Systems\\
University of Bonn\\
\texttt{rosu@ais.uni-bonn.de}\\
\And
Michael Weinmann\\
Computer Graphics Group\\
University of Bonn\\
\texttt{mw@cs.uni-bonn.de}\\
\And
Hassan Errami\\
Virtual Reality Group\\
University of Bonn\\
\texttt{errami@cs.uni-bonn.de}\\
\And
Sven Behnke\\
Autonomous Intelligent Systems\\
University of Bonn\\
\texttt{behnke@cs.uni-bonn.de}\\
\And
Maren Bennewitz\\
Humanoid Robots Lab\\
University of Bonn\\
\texttt{maren@cs.uni-bonn.de}\\
\And
Reinhard Klein\\
Computer Graphics Group\\
University of Bonn\\
\texttt{rk@cs.uni-bonn.de}\\
\And
Andreas Weber\\
Virtual Reality Group\\
University of Bonn\\
\texttt{weber@cs.uni-bonn.de}\\
\And
Angela Yao\\
National University of Singapore\\
\texttt{ayao@comp.nus.edu.sg}\\
\And
Juergen Gall\\
Computer Vision Group\\
University of Bonn\\
\texttt{gall@iai.uni-bonn.de}\\
}

\begin{document}
\maketitle

\begin{abstract}
The key prerequisite for accessing the huge potential of current machine learning techniques is the availability of large databases that capture the complex relations of interest.
Previous datasets are focused on either 3D scene representations with semantic information, tracking of multiple persons and recognition of their actions, or activity recognition of a single person in captured 3D environments.
We present Bonn Activity Maps, a large-scale dataset for human tracking, activity recognition and anticipation of multiple persons.
Our dataset comprises four different scenes that have been recorded by time-synchronized cameras each only capturing the scene partially, the reconstructed 3D models with semantic annotations, motion trajectories for individual people including 3D human poses as well as human activity annotations.
We utilize the annotations to generate activity likelihoods on the 3D models called activity maps.

\end{abstract}

% keywords can be removed
\keywords{Dataset \and Human Motion \and Tracking \and Recognition \and Activity \and Anticipation}

\section{Introduction}

Scene analysis, modeling and understanding are among the everlasting goals of computer vision, graphics and robotics.
Besides the impressive improvements resulting from sophisticated, task-specific learning approaches, the major progress achieved in the variety of different tasks is rooted in the development of datasets that capture the complex relationships between different scene aspects.
Examples include RGB-D datasets~\cite{firman2016rgbd}, datasets for human pose estimation and activity recognition~\cite{Joo_2015_ICCV,liu2017pku,koppula2013learning,h36m_pami,mhad,shahroudy2016ntu,Liu_2019_NTURGBD120,Kong_2019_ICCV,wu2015watch,spriggs2009temporal,chen2015utd,wang20143d,heilborn2015activitynet,lillo2014discriminative,li2016online,rahmani2016histogram}, material recognition~\cite{Dana:1997,hayman:2004,Li:2012:RMV:2404742.2404770,weinmann2014material,wang20164d,Bell:2013,bell2015material,Murmann_2019_ICCV}, semantic segmentation of RGB-D data~\cite{2017arXiv170201105A,song2015sun,Silberman:ECCV12} as well as 3D scenes~\cite{InteriorNet18,McCormac:etal:ICCV2017,fisher2012example,song2016ssc,Matterport3D,dai2017scannet,replica19arxiv,xiao2013sun3d,song2015sun,2017arXiv170201105A,scenenn-3dv16,uy-scanobjectnn-iccv19}, and the combination of geometry, semantics and activity maps~\cite{Savva:2014:scenegrok,Savva:2016:PLI}, i.e. likelihood maps of actions performed in certain scene regions with certain objects, to anticipate human behavior (see Table~\ref{tab:datasets}).
Our dataset differs from previous work~\cite{Savva:2014:scenegrok,Savva:2016:PLI} since it contains scenes with up to 12 visible persons at the same time.

\begin{figure*}
	\centering
	\includegraphics[width=\textwidth]{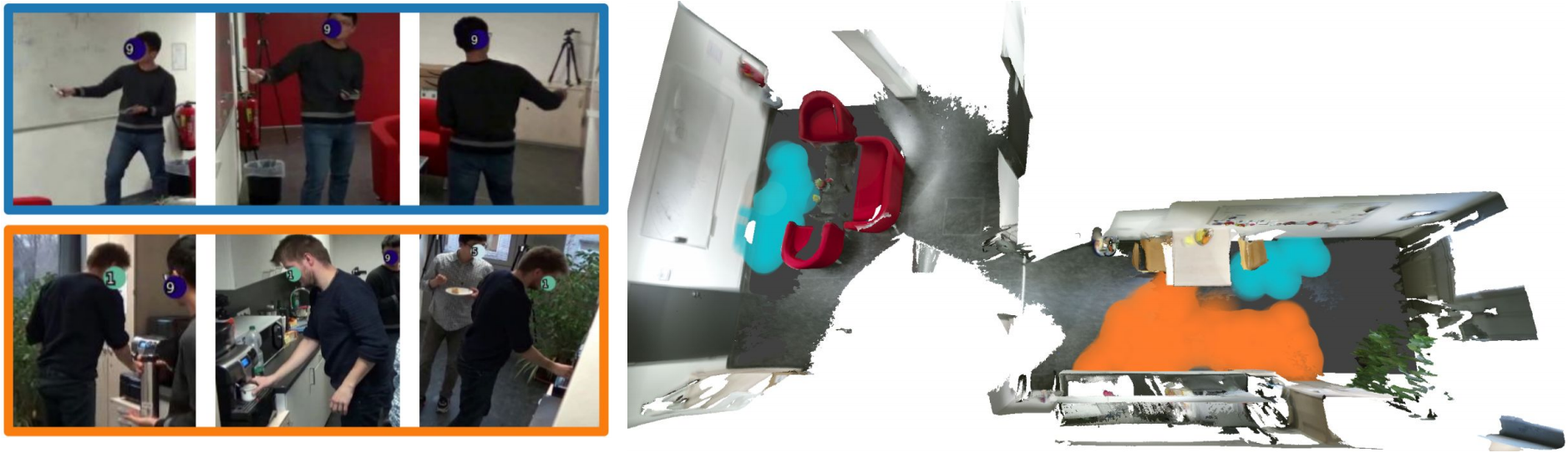}
	\caption{
	Bonn Activity Maps: Likelihood maps on the 3D model of two exemplary activities \textit{draw on whiteboard} (blue) and \textit{make coffee} (orange) are shown here. We annotate activities for every camera view, where each person is assigned a unique id (left), and generate activity maps for each 3D model indicating the locality of the respective activity over the entire recording and over all actors (right).
	}
	\label{fig:teaser1}
\end{figure*}

\begin{figure*}
	\centering
	\includegraphics[width=.70\textwidth]{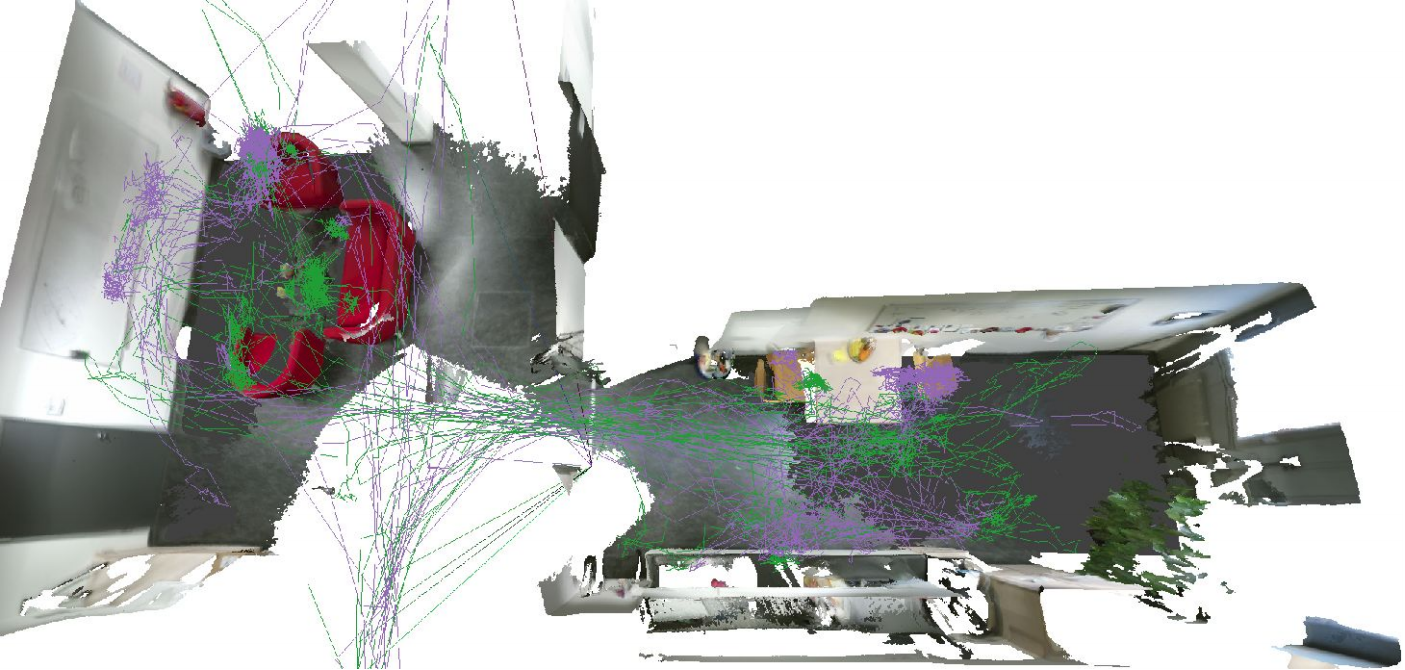}
	\caption{
	Bonn Activity Maps: 
	Trajectories of two actors are plotted into the 3D environment over the entire recording session.
	}
	\label{fig:teaser2}
\end{figure*}

To further foster research in this domain, we introduce Bonn Activity Maps. 
This large-scale dataset combines various types of scene information at the example of kitchen scenarios and their adjacent common areas (see Figures~\ref{fig:teaser1} and~\ref{fig:teaser2}).
In addition to multi-view scene video observation by pre-calibrated, time-synchronized cameras and accurate 3D models of the underlying scene geometry with semantic annotations, we additionally provide annotations regarding the poses, motion and activities of humans in the respective scenes, where each person is assigned a unique id which will remain consistent with that person, even in the event of leaving the scene for multiple minutes.
Each person can be assigned multiple activity labels, such as \textit{eat cake}, \textit{make coffee} or \textit{use smartphone}.
Since these labels are attached to a 3D object in the scene, they may transition between various camera views due to the fact that each camera only covers a small area in the recording volume.

\begin{table}[t]
    \fontsize{8.5pt}{10pt}\selectfont
    \centering
    \caption{Overview of recent datasets for 3D scene and human activity understanding. While many datasets target only one of these aspects, there are very few that contain both types of scene information. The Bonn Activity Maps dataset is the only dataset that contains 3D data of the environment as well as activities and tracked poses of multiple persons.}
    \label{tab:datasets}
    \begin{tabular}{lcccccc}
        \toprule
        Dataset &
        Camera Setup &
        3D Environment Data &
        Activities &
        Max Persons &
        Human Motion &
        Year\\
        \midrule
        Matterport3D~\cite{Matterport3D} & 1 RGB-D & 90 scenes & - & - & - & 2017 \\
        ScanNet~\cite{dai2017scannet} & 1 RGB-D & 707 scenes & - & - & - & 2017 \\
        Replica~\cite{replica19arxiv} & 1 RGB-D & 18 scenes & - & - & - & 2019 \\
        \midrule
        Berkeley MHAD~\cite{mhad} & 12 RGB, 2 RGB-D & - & 11 & 1 & 660 instances & 2013 \\
        CAD-120~\cite{koppula2013learning} & 1 RGB-D & - & 10 & 1 & \textapprox 1,200 instances & 2013 \\
        Panoptic Studio~\cite{Joo_2015_ICCV} & 480 RGB, 5 RGB-D & - & 5 & 8 & \textapprox 50 instances & 2015 \\
        NTU RGB+D~\cite{shahroudy2016ntu} & 3 RGB-D (17 configs) & - & 60 & 2 & 56,880 instances & 2016 \\
        PKU-MMD~\cite{liu2017pku} & 3 RGB-D & - & 51 & 2 & 21545 instances & 2017 \\
        NTU RGB+D 120~\cite{Liu_2019_NTURGBD120} & 3 RGB-D (32 configs) & - & 120 & 2 & 114,480 instances & 2019 \\
        MMAct~\cite{Kong_2019_ICCV} & 4 RGB + 1 Ego. RGB & - & 37 & 1 & 36,764 instances & 2019 \\
        \midrule
        SceneGrok~\cite{Savva:2014:scenegrok} & 1 RGB-D & 14 scenes & 7 & 1 & 1:51h video & 2014 \\
        PiGraphs~\cite{Savva:2016:PLI} & 1 RGB-D & 30 scenes & 43 & 1 & \textapprox 2:00h video & 2016 \\
        \midrule
        Bonn Activity Maps & 12 RGB, 1 RGB-D & 4 scenes & 60 & 12 & 5:22h video & 2019 \\
        \bottomrule
    \end{tabular}
\end{table}

In the following, we describe the data formats and how they are organized within the dataset (see Section~\ref{sec:dataset}) as well as the acquisition process (see Section~\ref{sec:dataaq}).
Please contact \texttt{bonn\_activity\_maps@iai.uni-bonn.de} for additional information or questions with regards to the dataset.
More information about the dataset can be found on \texttt{https://github.com/bonn-activity-maps/bonn\_activity\_maps}.

\section{Dataset}
\label{sec:dataset}

In this section, we describe Bonn Activity Maps and how the various data formats are organized within the dataset.
It consists of recordings of four different kitchens and their surrounding environment at the University of Bonn.
Figure~\ref{fig:SemPred} shows the 3D models of all four kitchens.
The kitchens share common characteristics such as similar furniture, floor, and walls but provide very different scales and geometries.
For each recording, we provide a folder with the following structure:
\begin{itemize}[leftmargin=*]
    \item \datafile{videos/}: Contains camera recordings and their associated calibrations, described in Section~\ref{sec:videorec}.
    \item \datafile{trajectories/}: Contains the human-annotated 3D trajectories of all individuals as well as their activity labels, described in Section~\ref{sec:trajectory}.
    \item \datafile{models3d/}: Contains the 3D models of the kitchens as well as 3D semantic segmentations of the data, described in Section~\ref{sec:3drecordings} and Section~\ref{sec:3dsegm}.
\end{itemize}
In the following, we will describe each of the aspects in detail.

% =============================================
% V I D E O  R E C O R D I N G S
% =============================================
\subsection{Video Recordings}
\label{sec:videorec}

One of the key aspects of the dataset are the captured activities in a kitchen environment.
Each of the four chosen kitchens were recorded for about two hours by 12 Sony HDR-PJ410 cameras -- the exact recording lengths are \texttt{01:26} hours, \texttt{01:59} hours, \texttt{01:57} hours and \texttt{02:00} hours.
The cameras are placed at fixed positions and record with a resolution of $ 1280 \times 720 $ at 25Hz.
For each camera, we provide an MP4 file which is time-synchronized with all other cameras as well as a file \texttt{calibration.json} which contains the extrinsic and intrinsic camera parameters.
In total, every camera recorded over 660,000 frames resulting in over 7,900,000 single frames when summing over all views and all recordings.
The 3D recording volume is spanned by the area that is covered by at least two cameras.
In detail, we provide the following files which are enumerated for each of the 12 cameras.

\begin{itemize}[leftmargin=*]
\item \datafile{videos/}
\begin{itemize}
    \item \datafile{camera00/}
    \begin{itemize}
        \item \datafile{calibration.json}: List of ranges of camera parameters defined as follows:
        \begin{itemize}
            \item \dataitem{start\_frame}: Starting frame from which this calibration is valid, with 0 being the first frame.
            The frame number corresponds to the frames in the recorded 25Hz video.
            \item \dataitem{end\_frame}: Last frame for which this calibration is valid.
            \item \dataitem{w}: Width of the video in pixels.
            \item \dataitem{h}: Height of the video in pixels.
            \item \dataitem{tvec}: 3D vector that transforms the global 3D camera position to the local position in camera space (origin). Part of the extrinsic calibration.
            \item \dataitem{rvec}: 3D vector that represents the cameras rotation in axis-angle format and transforms the global 3D camera orientation to the local orientation in camera space (z-direction). Part of the extrinsic calibration.
            \item \dataitem{K}: $ 3 \times 3 $ intrinsic camera matrix containing the focal lengths, the principal point, as well as the skewness parameter.
            \item \dataitem{distCoef}: Distortion coefficients in the order $ k_1, k_2, p_1, p_2, k_3 $.
        \end{itemize}
        \item \datafile{recording.mp4}: The recorded video data of the kitchen stored with a resolution of $ 1280 \times 720 $ at 25Hz in MP4 format.
    \end{itemize}
    \item \datafile{camera01/}
    \item $\hdots$
    \item \datafile{camera11/}
\end{itemize}
\end{itemize}

We use a list of ranges for \textdatafile{calibration.json} rather then a single entry since in one recording a camera was accidentally hit.
We thus provide two calibrations for this camera, one before and one after the hit occurred.

During the recording the number of persons in the camera views fluctuates strongly between only a single person to multiple people in close proximity overlapping each other.
Some of the actors were given tasks such as preparing coffee or having a phone call but the order in which to execute the task as well as the time could be chosen freely.
Furthermore, people were encouraged to participate in group activities or just follow their normal routines during the recording.
In comparison to other works, this results in very natural behavior.
This was further improved by the long recording time, relative to other datasets, as it made people \textit{get used} to the cameras.

% =============================================
% P E R S O N  T R A J E C T O R I E S
% =============================================
\subsection{Person Trajectories}
\label{sec:trajectory}

\begin{figure}[t]
    \centering
    \includegraphics[width=.8\linewidth]{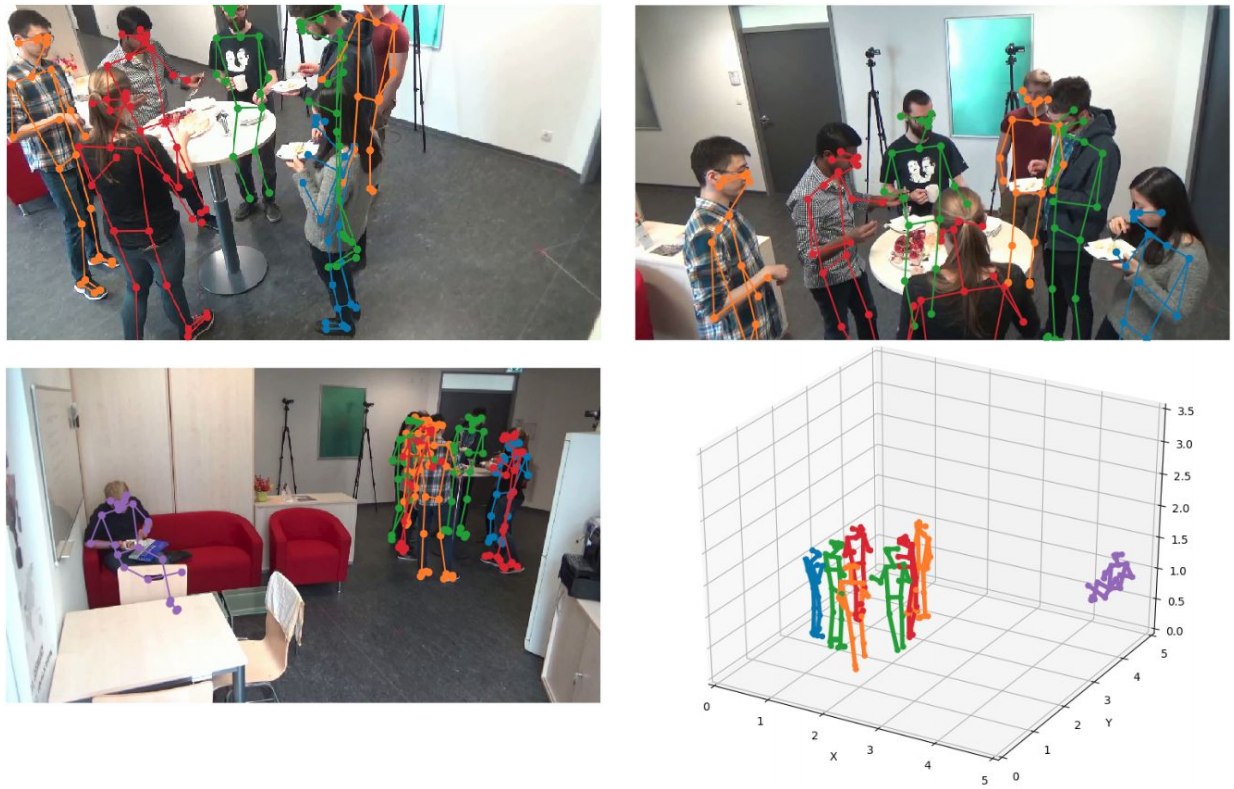}
    \caption{
    Generated 3D skeletons of persons (bottom right) projected into three camera views.
    }
    \label{fig:skel}
\end{figure}

A further crucial aspect of the dataset are the provided 3D person trajectories.
For each person that enters the recording volume a unique id is created.
Persons can enter and leave the scene but their unique id stays the same during the complete recording.
In case a person appears in multiple recordings, the unique id is kept to ensure cross-recording consistency.
Furthermore, a 3D point is annotated in each frame where a person is in the recording volume.
In addition, we provide 3D skeletons for each person which were generated using the method by Tanke and Gall~\cite{tanke2019iterative}.
The skeletons utilize the COCO keypoint challenge structure and 6 additional joints for the feet~\cite{cao2018openpose}.
An example of the generated skeleton data in a crowded scene is shown in Figure~\ref{fig:skel}.
The number of total persons in a recording vary from 18 to 30 persons with 12 being the maximum number of persons seen at the same time in a scene.
Furthermore, a total of 60 high-level activities were annotated during recording such as \textit{eat cake}, \textit{make coffee} or \textit{use smartphone}.
We provide a separate JSON file for each person which contains the following items:
\begin{itemize}[leftmargin=*]
\item \datafile{trajectories/}
\begin{itemize}
    \item \datafile{person000.json}
    \begin{itemize}
        \item \dataitem{frames}: Ordered list of all the frames where the person is visible during the recording. The frames correspond one-to-one to the frames of the 25Hz videos \textdatafile{recording.mp4}, described in Section~\ref{sec:videorec}.
        \item \dataitem{activities}: List of activities that the person is doing. This list has a one-to-one mapping to the \texttt{frames} property, but a person can do multiple activities at the same time. The list of all activities is described in Table~\ref{tab:labels}.
        \item \dataitem{positions}: List of 3D points, representing the 3D location of the person. This list has a one-to-one mapping to the
        \dataitem{frames} property and is human-annotated.
        \item \dataitem{poses}: List of 3D poses, representing the 3D pose of the person. This list has a one-to-one mapping to the
        \dataitem{frames} property and is generated using the method by Tanke and Gall~\cite{tanke2019iterative}. A pose consists of 24 joints each being represented by a 3D point.
    \end{itemize}
    \item \datafile{person001.json}
    \item $\hdots$
\end{itemize}
\end{itemize}
\begin{table}
\fontsize{8.5pt}{10pt}\selectfont
\centering
\caption{List of the 60 annotated activities.}
\label{tab:labels}
\begin{tabular}{ccc}
    \toprule
    Carry cake &
    Carry cup &
    Carry kettle \\
    Carry milk &
    Carry plate &
    Carry whiteboard eraser \\
    Carry whiteboard marker &
    Check water in coffee machine &
    Clean countertop \\
    Clean dish &
    Close cupboard &
    Close dishwasher \\
    Close drawer &
    Close fridge &
    Cut cake \\
    Draw on whiteboard &
    Drink &
    Eat cake \\
    Eat fruit &
    Empty cup in sink &
    Empty ground coffee \\
    Empty water from coffee machine &
    Erase on whiteboard &
    Fill coffee beans \\
    Fill coffee water tank &
    Make coffee &
    Mark coffee in list\\
    Open cupboard &
    Open dishwasher &
    Open drawer \\
    Open fridge &
    Peel fruit &
    Place cake on plate \\
    Place cake on table &
    Place cup onto coffee machine &
    Place in microwave \\
    Place sheet onto whiteboard &
    Place water tank into coffee machine &
    Pour kettle \\
    Pour milk &
    Press coffee button &
    Put cake in fridge \\
    Put cup in microwave &
    Put in dishwasher &
    Put sugar in cup \\
    Put teabag in cup &
    Put water in kettle &
    Read paper \\
    Remove sheet from whiteboard &
    Start dishwasher &
    Start microwave \\
    Take cake out of fridge &
    Take cup from coffee machine &
    Take out of microwave \\
    Take teabag &
    Take water from sink &
    Take water tank from coffee machine \\
    Use laptop &
    Use smartphone &
    Wash hands \\
    \bottomrule
\end{tabular}
\end{table}

% =============================================
% 3 D  D A T A
% =============================================
\subsection{3D Data Recordings}
\label{sec:3drecordings}

In addition to the videos of recorded activities as well as person trajectory data, we also provide RGB-D and reconstructed 3D mesh data of the kitchen environments.
For this, we used a Microsoft Kinect v2 sensor which records RGB data with a native resolution of $ 1920 \times 1080 $ as well as depth images with a resolution of $ 512 \times 424 $ both at 30Hz.
These input streams are already time-synchronized by the sensor's firmware and cover all relevant objects and scene content.
Across the four kitchen scenes, we recorded RGB-D image data which were used to reconstruct high-resolution reference 3D meshes.
In particular, we provide the following data:
\begin{itemize}[leftmargin=*]
\item \datafile{models3d/}
\begin{itemize}
    \item \datafile{kitchen00/}
    \begin{itemize}
        \item \datafile{input\_rgbd\_stream/}: Recorded RGB-D image data in TUM RGB-D dataset~\cite{sturm2012benchmark} format:
        \begin{itemize}
            \item \datafile{rgb/}: Directory containing the recorded color data stored as 3-channel 8-bit PNG images.
            \item \datafile{depth/}: Directory containing the recorded depth data stored as 1-channel 16-bit PNG images. Depth values measured in meter are scaled by a factor of 5000.
            \item \datafile{rgb.txt}: Order of the RGB image sequence.
            \item \datafile{depth.txt}: Order of the depth image sequence. These two sequences are already time-synchronized such that the $ i $-th image in both text files corresponds to the same point in time during recording.
        \end{itemize}
        \item \datafile{intrinsics.json}: Intrinsic camera parameters of the RGB-D sensor
        \begin{itemize}
            \item \dataitem{depth}: Intrinsic parameter of the depth camera, i.e. \dataitem{w}, \dataitem{h}, \dataitem{K}, and \dataitem{distCoef} (see~Section~\ref{sec:videorec})
            \item \dataitem{rgb}: Intrinsic parameters of the RGB camera. Same format as for \dataitem{depth}.
            \item \dataitem{tvec}: 3D translation vector between both cameras.
            \item \dataitem{rvec}: 3D rotation vector between both cameras in axis-angle format. The resulting rigid transformation maps a 3D point from the depth camera coordinate system to the one of the RGB camera.
        \end{itemize}
        \item \datafile{initial\_camera\_pose.json}: Initial 6-DOF camera pose relative to the calibration coordinate system~(see Section~\ref{sec:calibration}) in JSON format.
        \begin{itemize}
            \item \dataitem{tvec}: 3D translation vector to the global coordinate system.
            \item \dataitem{rvec}: 3D rotation vector to the global coordinate system in axis-angle format.
        \end{itemize}
        \item \datafile{reference\_output\_mesh.ply}: Reference 3D mesh reconstructed from the RGB-D data and stored in PLY format.
    \end{itemize}
    \item \datafile{kitchen01/}
    \item $\hdots$
    \item \datafile{kitchen03/}
\end{itemize}
\end{itemize}
To improve the usability and accessibility of the dataset, we adopt widely used file formats for data storage.
For the RGB-D image sequences, we use the same format as for the TUM RGB-D dataset~\cite{sturm2012benchmark} which is commonly known in the computer vision and robotics communities.
Similar to video data (see Section~\ref{sec:videorec}), we have chosen a simple format for the intrinsic camera parameters based on the OpenCV calibration toolkit API.
Note that we provide the raw image data without any further pre-processing (besides the internal time-synchronization) to reduce the probability of introducing biases into the data.
Therefore, we also provide the 6-DOF rigid transformation between the depth and RGB camera which is required for camera calibration.

% =============================================
% 3 D  D A T A S E G
% =============================================
\subsection{3D Data Segmentation}
\label{sec:3dsegm}

%%% the 4 kitchens
\bgroup
\def\ElemWidth{3.7cm}
\def\Img{./img/sem_pred/k1_rgb.png} 
\newlength{\WS}
\newlength{\HS}
\settowidth{\WS}{\includegraphics{\Img}}
\settoheight{\HS}{\includegraphics{\Img}}
\begin{figure*}
	\centering
	\begin{tabular}{cccc}
		\includegraphics[trim=.29\WS{} .14\HS{} .34\WS{} .16\HS{},clip, width=\ElemWidth{}]{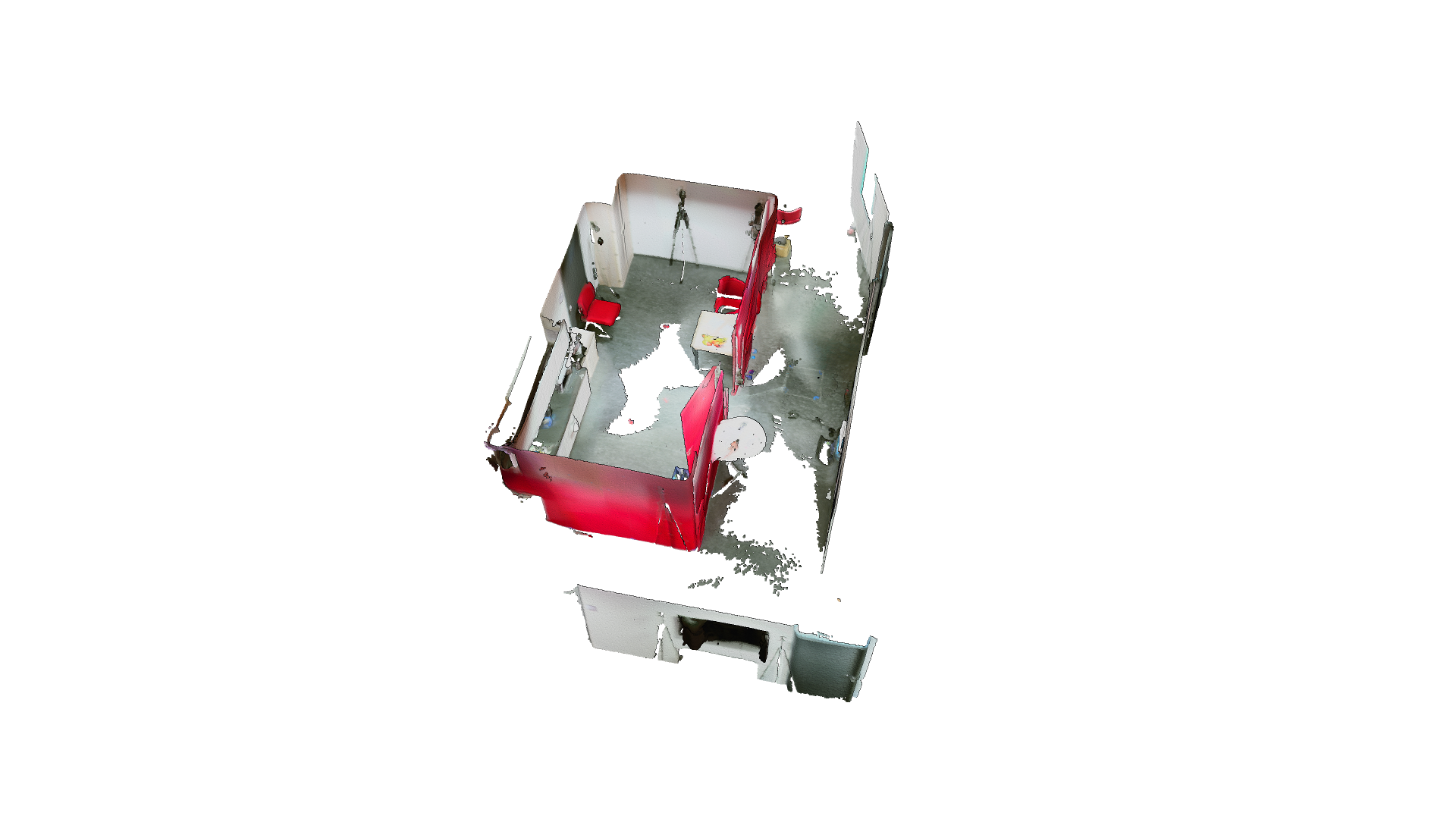}& 
		\includegraphics[trim=.33\WS{} .21\HS{} .37\WS{} .16\HS{},clip, width=\ElemWidth{}]{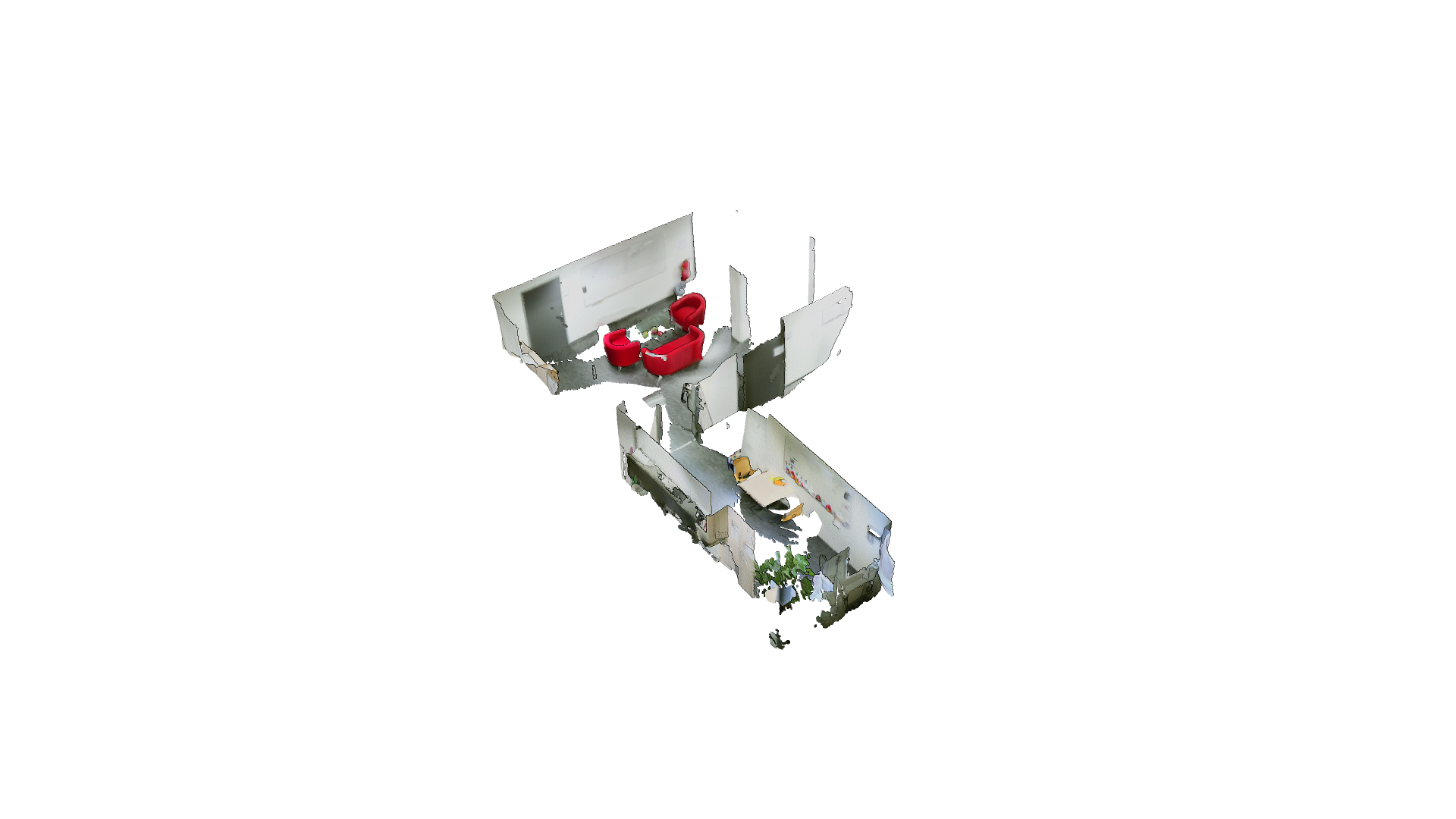}& 
		\includegraphics[trim=.29\WS{} .15\HS{} .32\WS{} .16\HS{},clip, width=\ElemWidth{}]{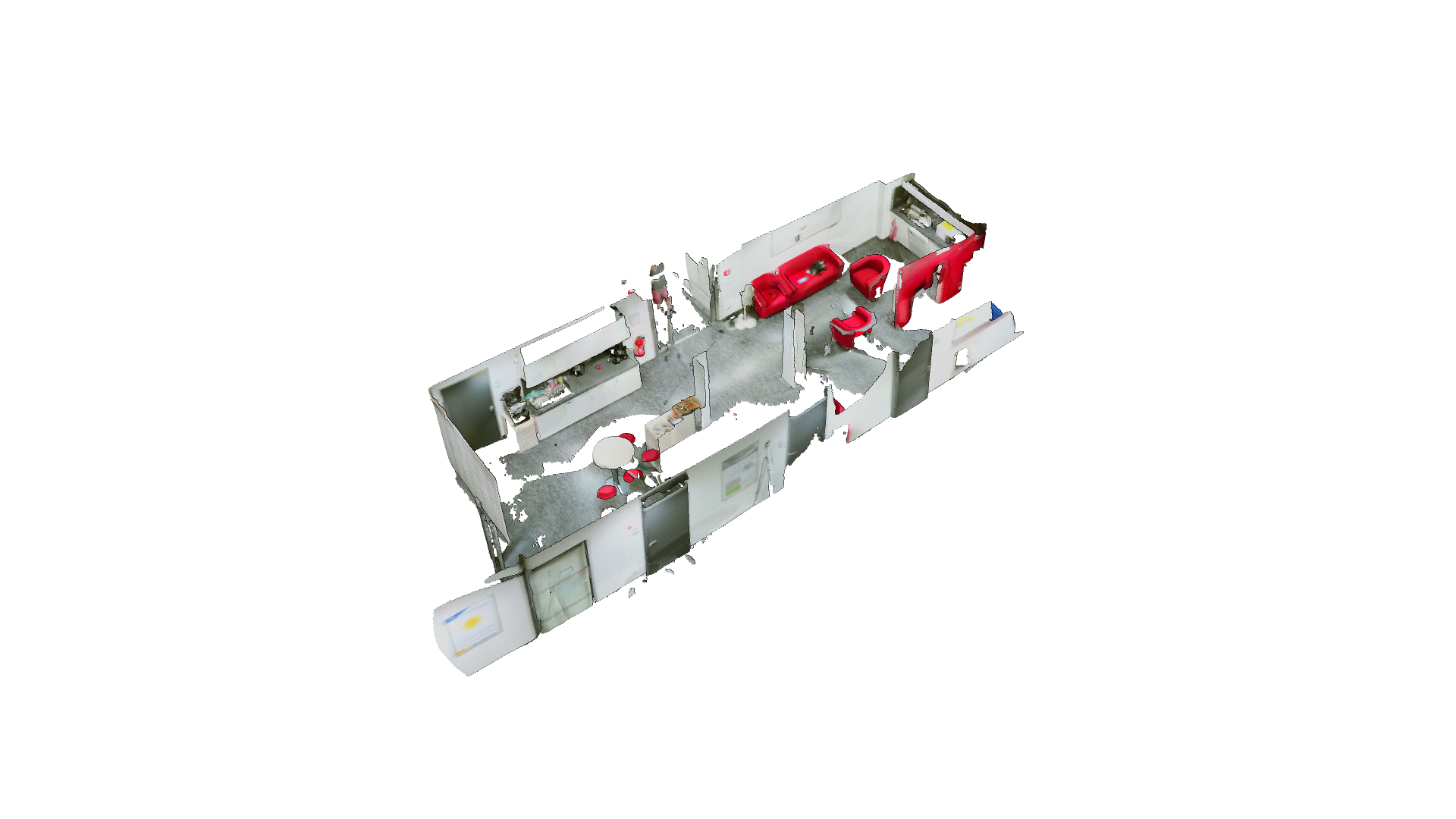}& 
		\includegraphics[trim=.33\WS{}  .2\HS{} .30\WS{} .16\HS{},clip, width=\ElemWidth{}]{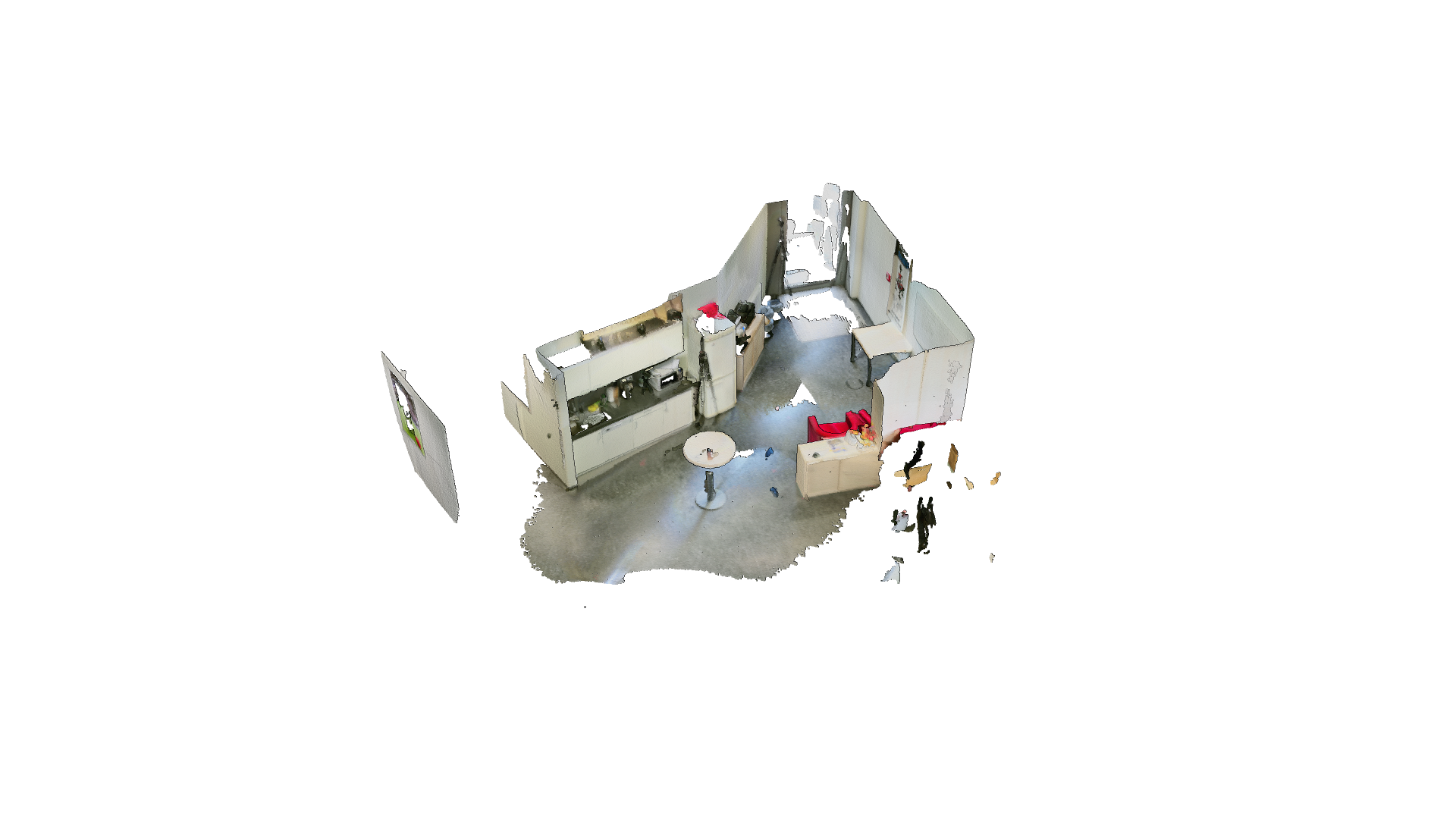} 
		\\ 
		\includegraphics[trim=.29\WS{} .14\HS{} .34\WS{} .16\HS{},clip, width=\ElemWidth{}]{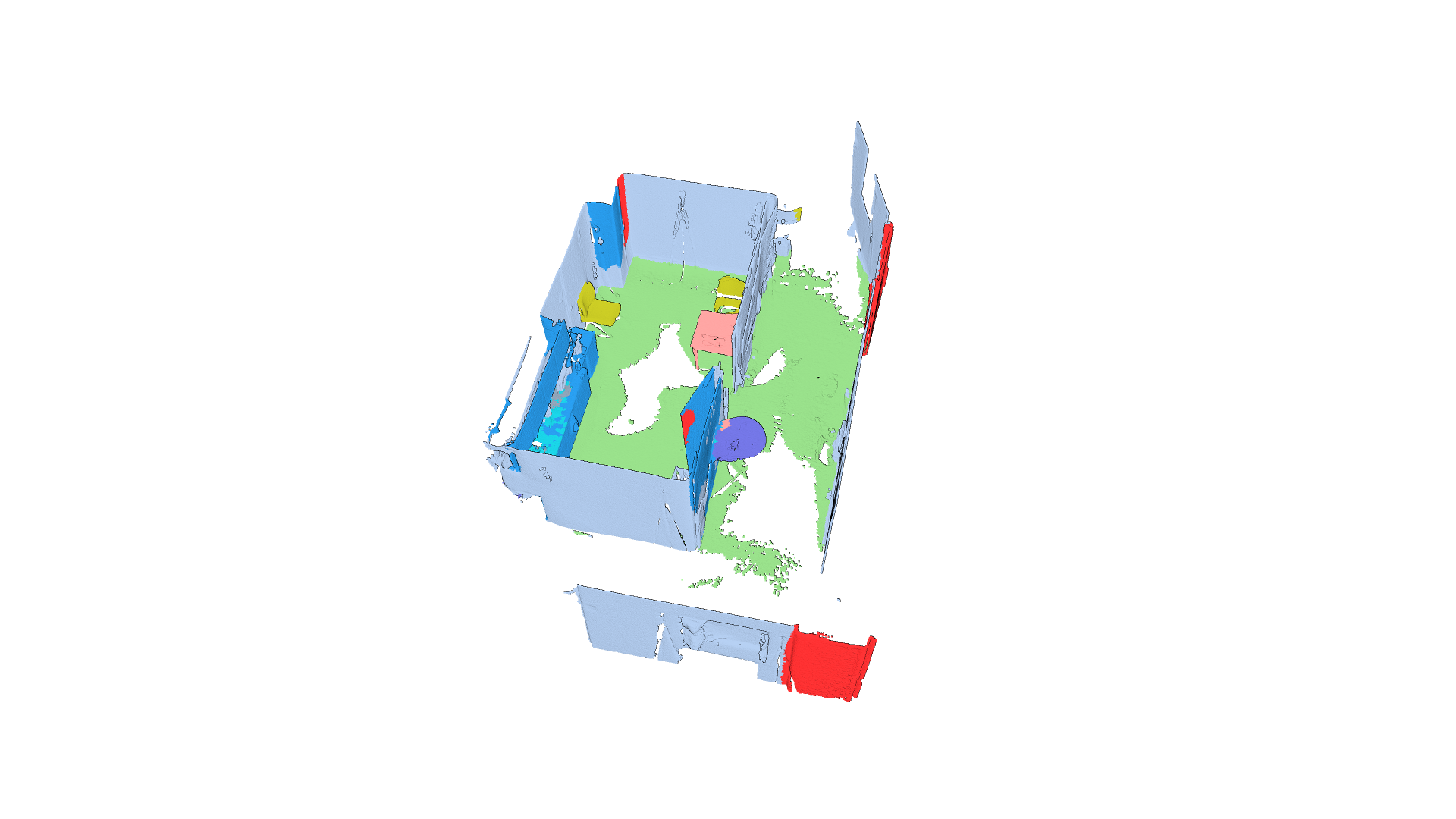}& 
		\includegraphics[trim=.33\WS{} .21\HS{} .37\WS{} .16\HS{},clip, width=\ElemWidth{}]{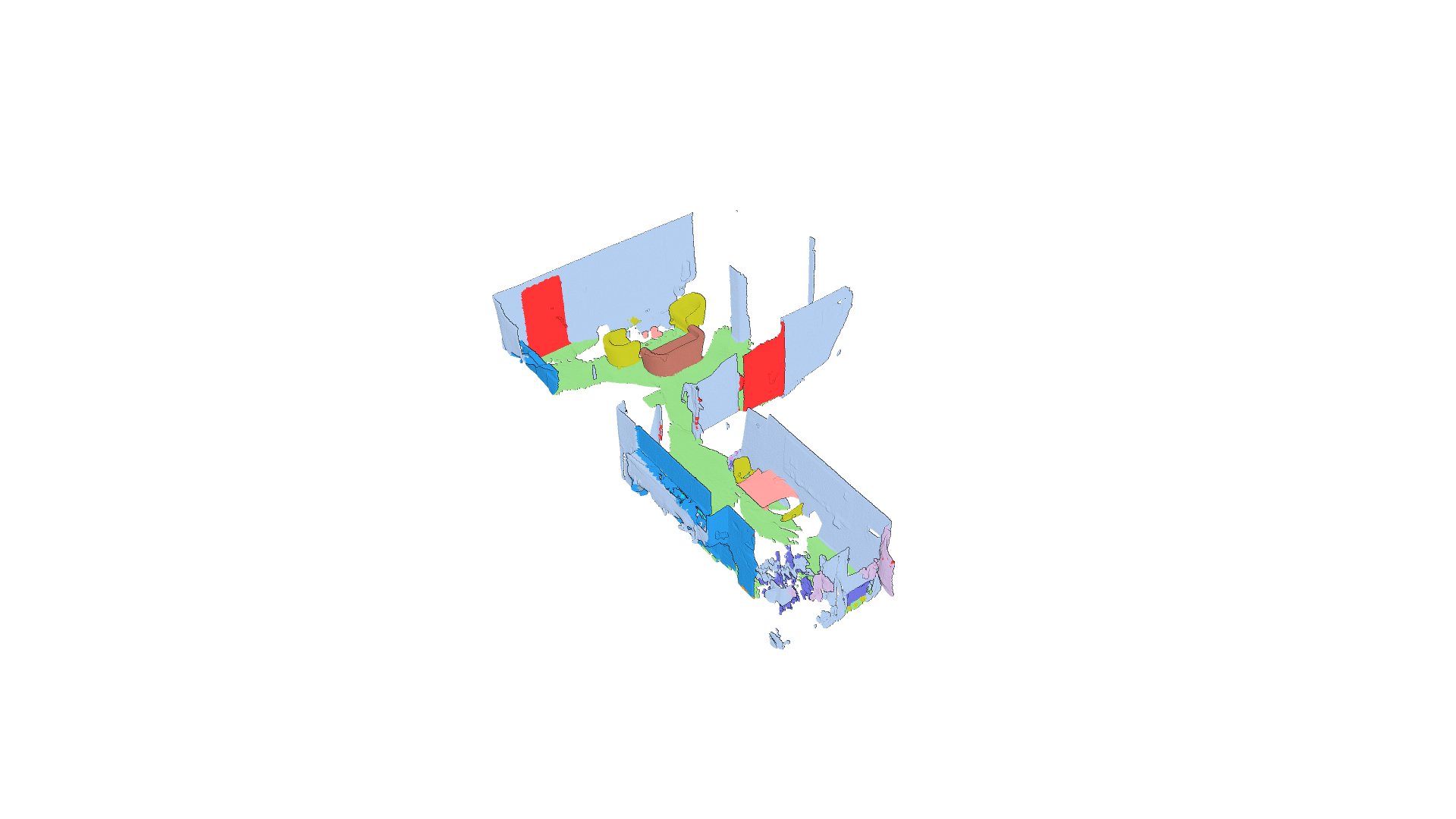}& 
		\includegraphics[trim=.29\WS{} .15\HS{} .32\WS{} .16\HS{},clip, width=\ElemWidth{}]{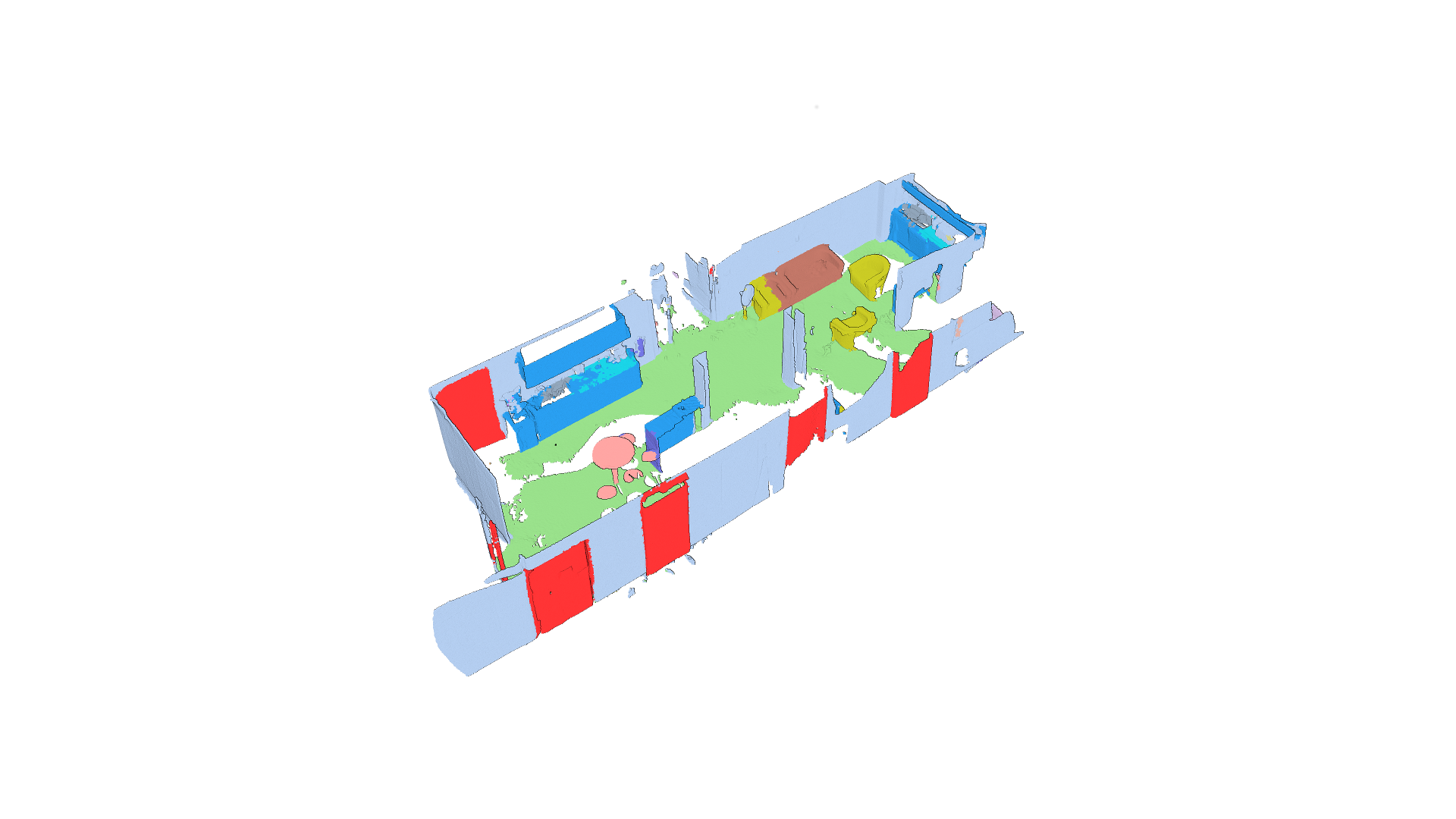}& 
		\includegraphics[trim=.33\WS{}  .2\HS{} .30\WS{} .16\HS{},clip, width=\ElemWidth{}]{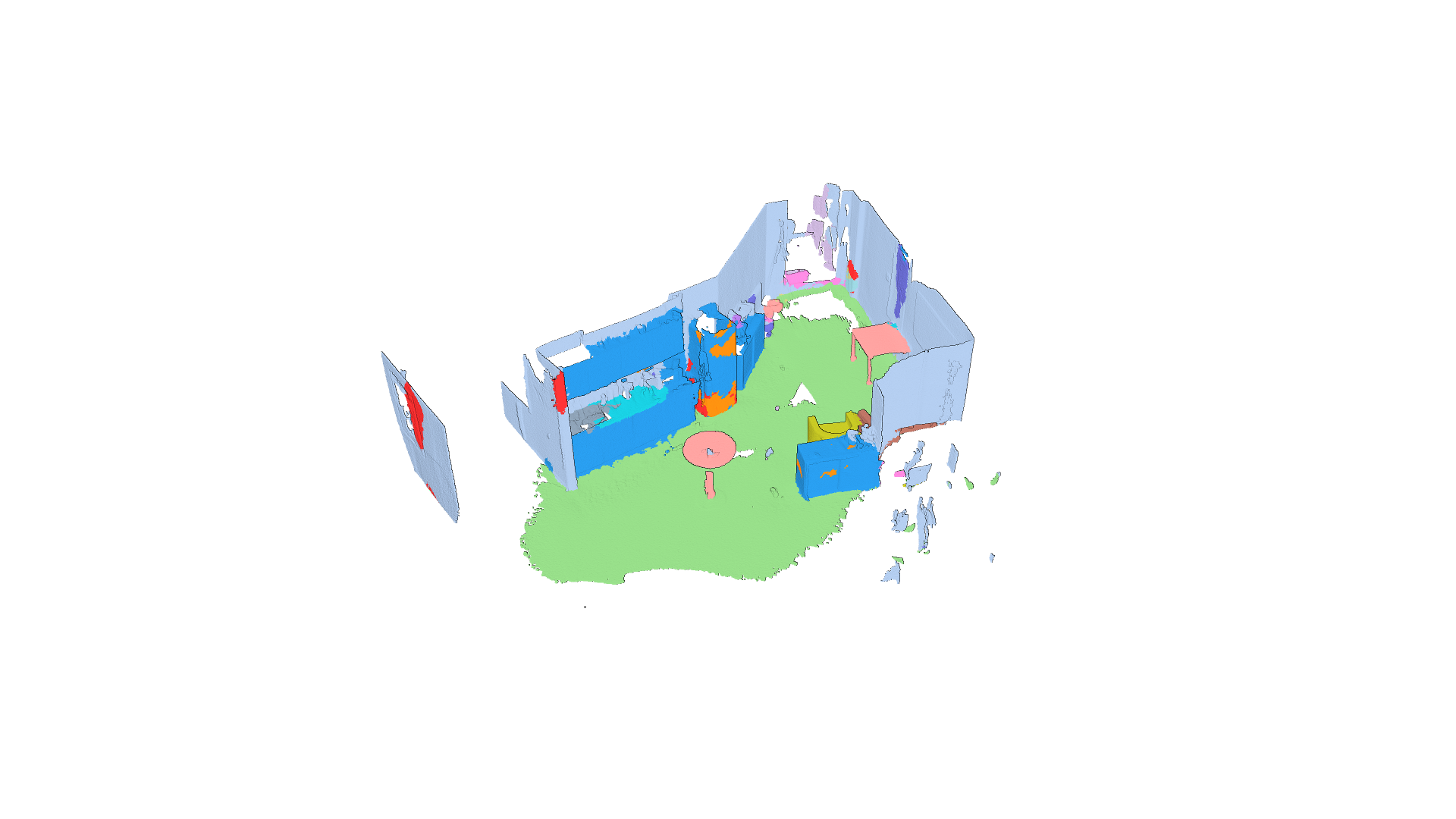}  
	\end{tabular}
	\caption{Semantic predictions: Colored reference meshes are reconstructed and provided as well as their semantically segmented counterparts where the color coding of the semantic labels corresponds to the ScanNet dataset~\cite{dai2017scannet}.}
	\label{fig:SemPred}
\end{figure*}
\egroup

The reconstructed 3D meshes are used for prediction of semantic classes.
We trained LatticeNet~\cite{latticenet} on the ScanNet room dataset~\cite{dai2017scannet} and evaluate it on each kitchen mesh individually.
Since LatticeNet can deal with raw point clouds, we ignore the connectivity information of the faces and only use the vertices of the mesh as input.
As a result, we obtain per-vertex probability values over the 21 labels annotated in ScanNet, e.g. chairs, tables, windows, etc.
Therefore, we provide a further PLY file for each scene in which every 3D point has an additional attribute corresponding to the class label with the highest probability and an attribute for the respective color according to the ScanNet color scheme.
\begin{itemize}[leftmargin=*]
\item \datafile{models3d/}
\begin{itemize}
    \item \datafile{kitchen00/}
    \begin{itemize}
        \item \datafile{reference\_output\_mesh\_with\_labels.ply}: Reconstructed and additionally labelled reference 3D mesh stored in PLY format.
    \end{itemize}
    \item \datafile{kitchen01/}
    \item $\hdots$
    \item \datafile{kitchen03/}
\end{itemize}
\end{itemize}
Results of the 3D meshes and their corresponding class label predictions are shown in Figure~\ref{fig:SemPred}.

\section{Data Acquisition}
\label{sec:dataaq}

\begin{figure}[t]
    \centering
    \includegraphics[width=.49\linewidth]{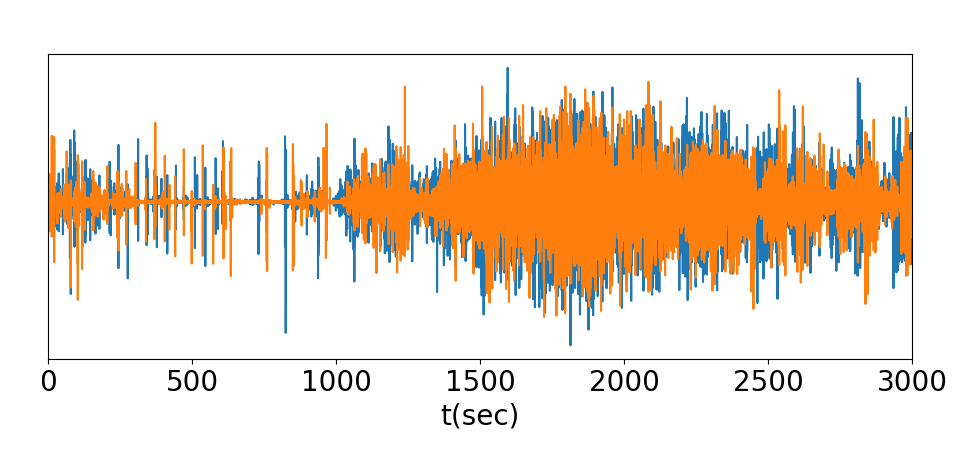}
    \includegraphics[width=.49\linewidth]{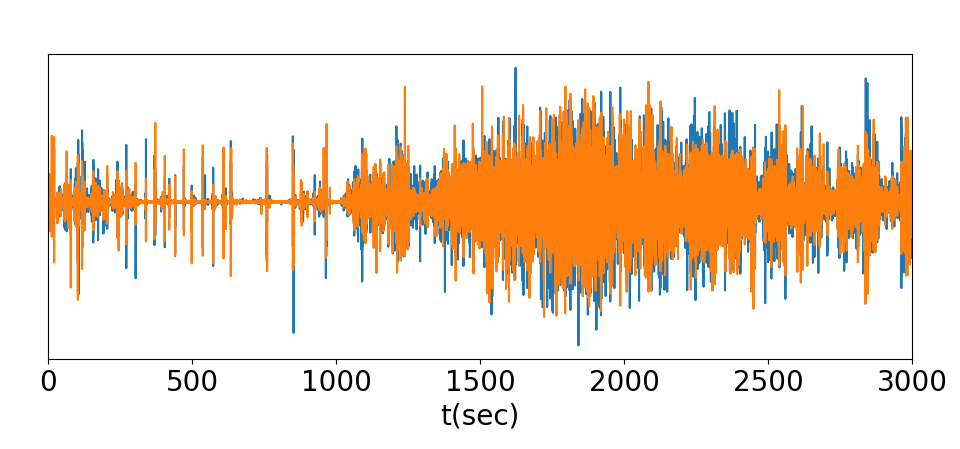}
    \caption{
Time synchronization of video recordings. We align the peaks of the unsynchronized videos (left) to ensure temporal consistency across different views.
    }
    \label{fig:s0}
\end{figure}

\begin{figure}[t]
    \centering
    \includegraphics[width=.4\linewidth]{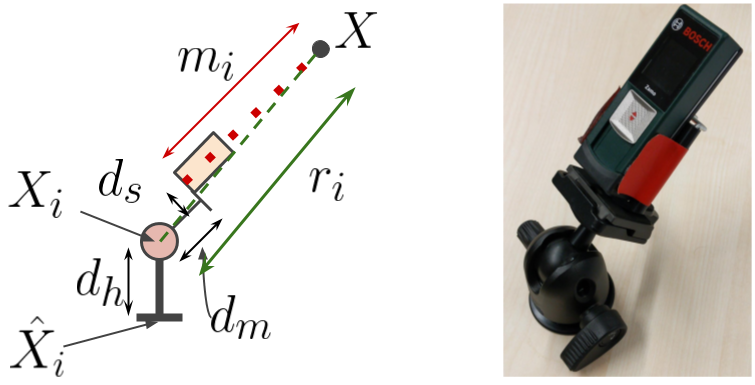}
    \caption{
Laser measuring device: We utilized a laser which we mounted onto a camera gimbal to measure out-of-plane distances (right).
In the schematic overview of the apparatus (left), $X_i$ denotes the center of the ball-bearing, $X$ is the target point, $m_i$ is the distance to $X$, measured by the laser, $d_s$ is the top offset of the laser from the rotation axis, while $d_m$ represents the forward offset.
$d_h$ is the distance of $X_i$ to the ground.
    }
    \label{fig:laser}
\end{figure}

\begin{figure}[t]
    \centering
    \includegraphics[width=.88\linewidth]{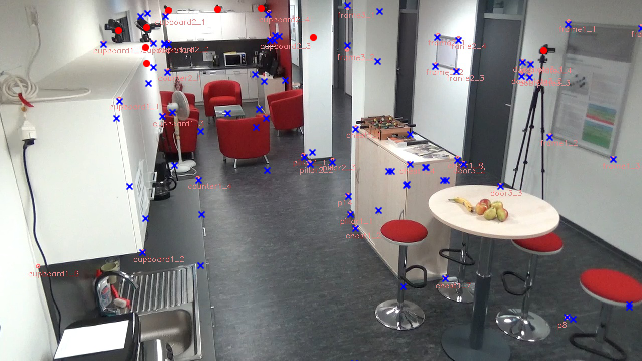}
    \caption{
Example frame of a calibrated camera.
Small light-red circles represent 2D annotations of the landmark points that were hand-clicked, blue crosses represent the 3D positions of the landmarks which are projected into the calibrated camera view, and large red points represent the 3D locations of the other cameras projected into the camera view.
Note that 3D positions in a camera view might be occluded by the scene.
    }
    \label{fig:calibration}
\end{figure}

In the following, we provide a description of the data acquisition process.

% ====================================================
% S Y N C H R O N I Z A T I O N
% ====================================================
\subsection{Synchronization}

When working in multi-camera setups, the synchronization of the videos is essential to ensure temporal consistency across views.
Unfortunately, the utilized off-the-shelf cameras do not provide hardware-synchronization and, hence, we resort to the following approach:
We record the videos independently and synchronize them afterwards in a post-processing step using the audio channel.
To aid the offline synchronisation, we produce sharp sounds using a clapperboard at the beginning and end of the recordings.
We can then synchronize two signals (videos) by finding the peak of the cross-correlation between them, which can be easily done using the Fourier transform:
\begin{equation}
    (f \star g)(\tau)
    := \int_{-\infty}^{\infty} f(t) \, g(t+\tau) \, dt
    = \int_{-\infty}^{\infty} f(t-\tau) \, g(t) \, dt
    = f(-\tau) * g(\tau)
    = \mathcal{F}^{-1}(F^{*} \cdot G)(\tau)
\end{equation}
After finding the common start and end point, we cut the videos accordingly.
Figure~\ref{fig:s0} shows an example of two videos before and after time synchronization.

% ====================================================
% C A L I B R A T I O N 
% ====================================================
\subsection{Calibration}
\label{sec:calibration}

To obtain the intrinsic camera parameters $K$, we utilize a checkerboard and the well-known algorithm described in the work of Zhang~\cite{zhang2000flexible}.

For extrinsic parameter calibration, we have to solve the Perspective-$n$-Point (P$n$P) problem~\cite{lepetit2009epnp} for which we need a set of 3D points in world coordinates and their respective 2D positions in the camera image.
We need to cover the complete recording volume with well-known 3D points since each camera only covers a small section.
To ease our annotation work, we use easy-to-recognize landmark points such as corners of cupboards, whiteboards and fire warning signs.
For cameras that face mostly uniform walls we create artificial landmarks using colored tape.

To place these landmark points into a joint global coordinate space, we follow a multi-stage approach since our scenes are non-convex and our measuring device only accurately measures in-plane distances.
We measure a set of base points $\hat{X}_i$ which all reside on the ground plane ($z=0$) to get an initial set of landmarks.
For this, we first pick a point on the ground to be the origin and then choose one point on the $x$- as well as one on the $y$-axis of our right-handed coordinate frame and measure the distance from the origin.
Finding positions of other points on the ground plane can now be solved by gradient descent using the known positions of the base points and their distance to the target point. 
We use an off-the-shelf range measuring laser to measure the actual distances.

To increase the set of landmarks to also cover arbitrary point $X$ in 3D space, we need the calculate the distances $r_i, r_j, r_k$ to at least three already known points $X_i, X_j, X_k$.
We obtain these distances by a laser measuring device that is mounted onto a camera gimbal as described in Figure~\ref{fig:laser}.
Here, we need to account for the gimbal to get the correct values:
\begin{equation}
    r_i = \sqrt{(m_i + d_m)^2 + d_s^2}
\end{equation}
$d_s$ and $d_m$ are offset constants given by the gimbal while $m_i$ is the laser-measured distance for the laser apparatus at point $X_i$ to target point $X$.
As the ball hinge is placed several millimeters above $z=0$, we also need to adjust for the height $d_h$:
\begin{equation}
    X_i = \hat{X}_i +
    \begin{bmatrix}
        0 & 0 & d_h
    \end{bmatrix}^T
\end{equation}
When we measure base points with $z=0$, we do not use the gimbal and the distance simplifies to $r_i = m_i$.
For each known point, we can now setup the constraint
\begin{equation}
    (x-x_{i})^2 + (y-y_{i})^2 + (z-z_{i})^2 = r_i^2
\end{equation}
with $X=[ x \ y \ z]$ and $X_i = [x_i \ y_i \ z_i ]$.
To get the target position $X$, we solve the system with least-squared error
\begin{equation}
    L(x, y, z) = \frac{1}{N} \sum_{i}^{N} \left\Vert \sqrt{(x-x_{i})^2 + (y-y_{i})^2 + (z-z_{i})^2 }  - r_{i} \right\Vert _2^2
\end{equation}
by gradient descent.
Figure~\ref{fig:calibration} shows an example of a calibrated kitchen with various landmark points.

% ====================================================
% 3 D  R E C O R D I N G  
% ====================================================
\subsection{3D Data Recordings}
\label{sec:dataaq3d}

\begin{figure}[t]
    \centering
    \includegraphics[width=.8\linewidth]{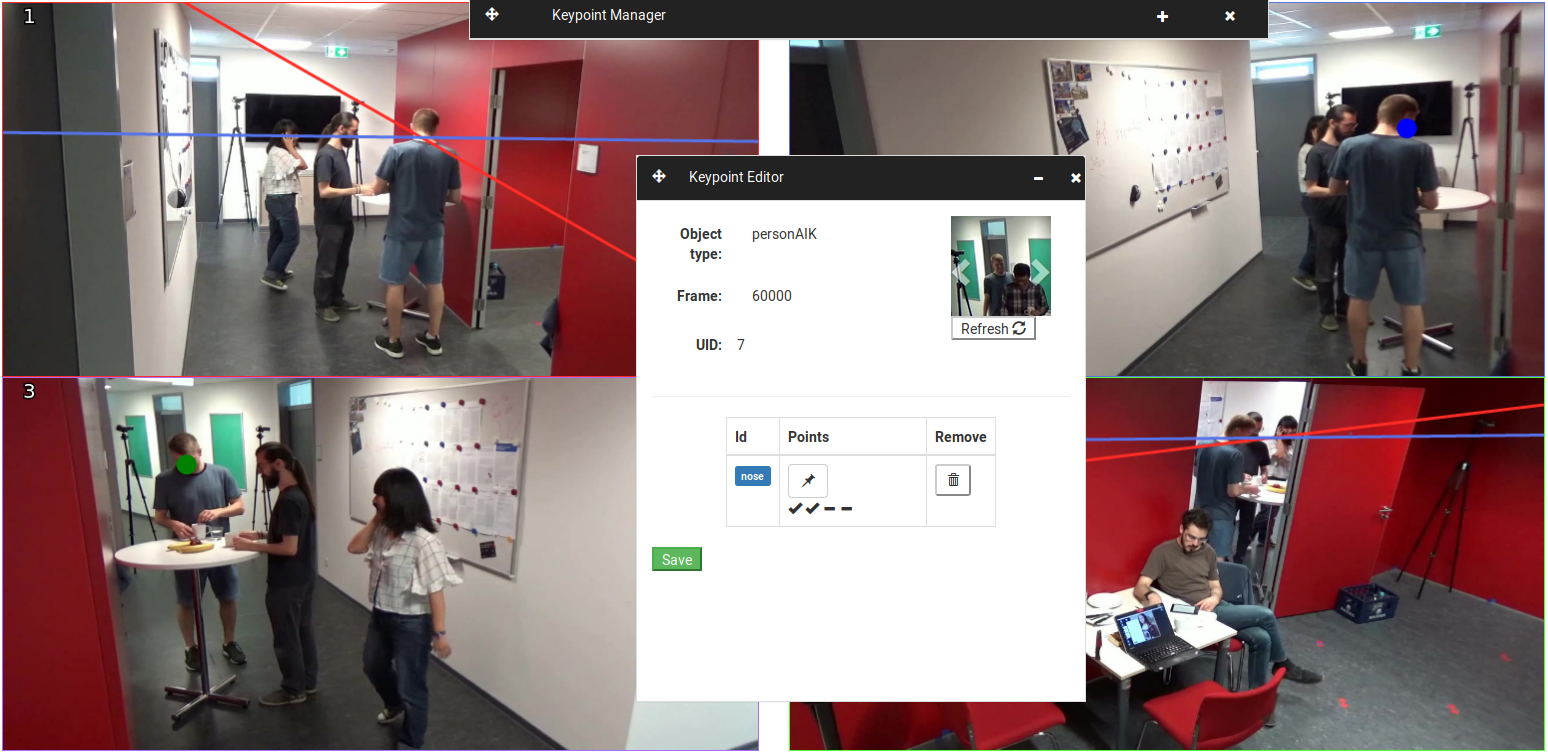}
    \caption{
Activity annotation tool.
The tools shows up to four camera views in which the a person can be annotated.
To create a 3D point, at least two views have to be utilized while additional views can be used to obtain more accurate results.
If a 2D point is selected in one or more views (top right and bottom left), epipolar lines are drawn in the other views to guide the annotator (top left and bottom right).
    }
    \label{fig:annotationtool}
\end{figure}

In addition to the video data, we also captured the scene using an RGB-D camera to obtain a 3D model of the scene geometry.
For this, we used a Microsoft Kinect v2 to record the raw RGB-D data.
In order to reconstruct 3D models from these input data, we use standard volumetric 3D reconstruction approach~\cite{niessner,infinitam,stotko2019groupscale}.
Although significant advances in 3D reconstruction have been achieved~\cite{Zollhoefer2018state}, the reconstruction accuracy may still not be sufficient for certain parts of the model, e.g. in regions with challenging illumination conditions or only very few geometric and photometric features.
Therefore, we provide both the captured RGB-D input streams including the intrinsic parameters of our camera as well as the reference 3D models.
This allows users to directly use the reference 3D model for a fair and comparable evaluation of their technique as well as to generate their own models with a different reconstruction approach to leverage improvements in terms of accuracy.
A common and widely used choice of the world coordinate system of the reconstructed model is the camera coordinate system at the first frame, i.e. choosing the identity as the initial camera pose.
Since the calibration coordinate system is set as the reference system for the whole dataset, we align the 3D model and provide the respective transformation as the coordinates of the initial camera pose.
Note that the RGB-D image acquisition was performed before the video recordings and the resulting 3D models represent the state of the kitchen before any action has been performed.

% ====================================================
% D A T A  A N N O T A T I O N  
% ====================================================
\subsection{Data Annotation}

\begin{figure}[t]
    \centering
    \includegraphics[width=.8\linewidth]{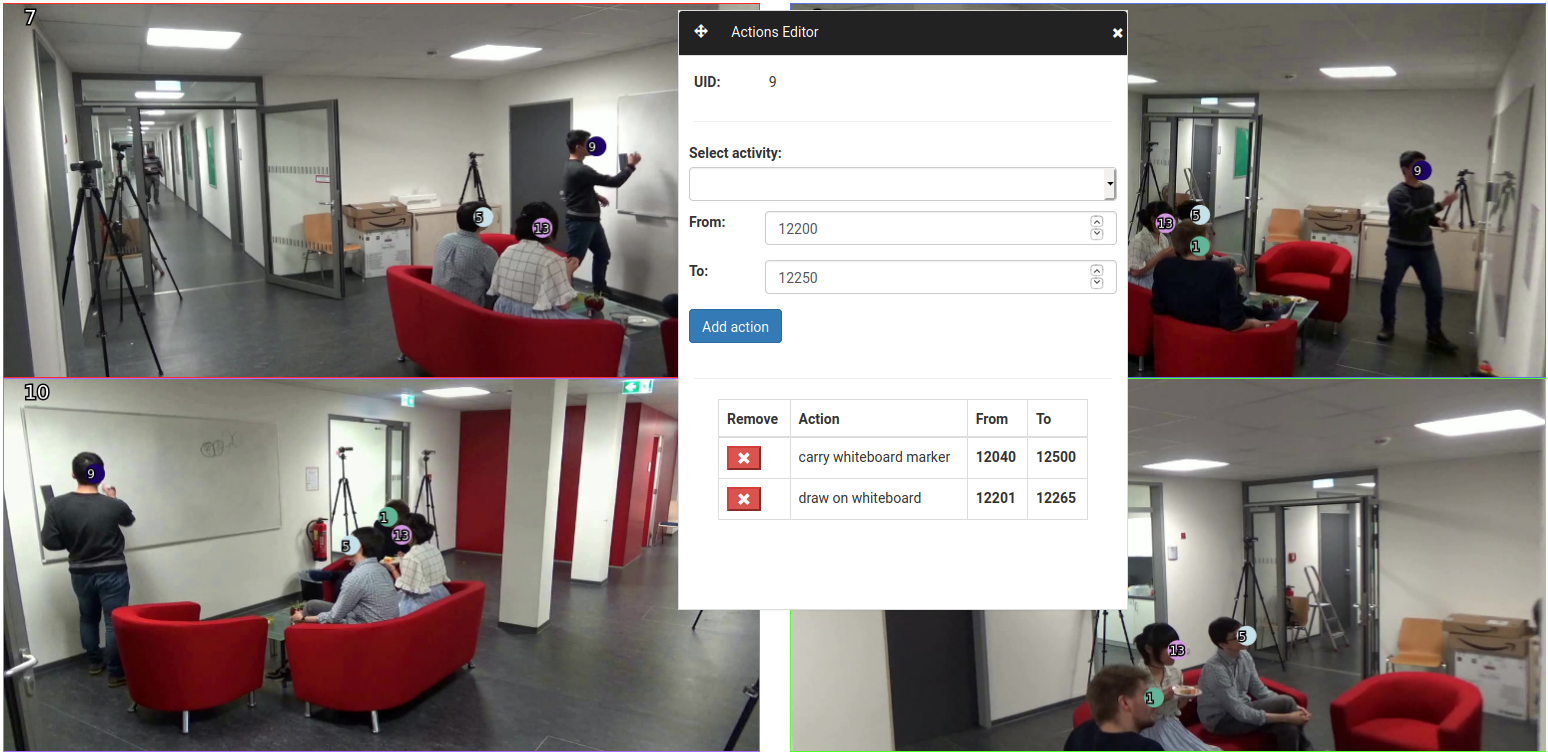}
    \caption{
Activity annotation tool.
An annotator can choose an activity as well as its starting and end point. Activities can overlap each other and are always associated to a person which is represented as a 3D point.
    }
    \label{fig:actionannot}
\end{figure}

For data annotation, we developed a new tool. 
Since each camera only covers a small portion of the recording volume, annotators must have easy access to other views to switch quickly between them.
For annotating 3D objects, at least two views have to be utilized to allow triangulation.
However, we allow to annotate points in up to four views which is shown in Figure~\ref{fig:annotationtool}.
For comfort and performance, the tool separates the videos into small chunks that can be easily annotated while retaining global information such as currently active actions and person ids.
Moving on to the next or previous chunk can be easily done with the tool.
Figure~\ref{fig:annotationtool} shows how 3D person ids are annotated while Figure~\ref{fig:actionannot} shows how activities are annotated.
As activities are always associated with a 3D person trajectory, we first had to completely annotate the trajectories before labeling activities.
The list of all 60 labeled activities can be seen in Table~\ref{tab:labels}.
Furthermore, an exemplary activity map as well as trajectories of two persons are shown in Figures~\ref{fig:teaser1} and \ref{fig:teaser2}, respectively.

\section{Summary}

We presented a description of Bonn Activity Maps, a large-scale dataset for human tracking, activity recognition and anticipation of multiple persons.
Our dataset combines various types of scene information at the example of kitchen scenarios where each recording contains time-synchronized video sequences of the scene, human-annotated 3D trajectories and generated 3D human poses of all individuals as well as their activity labels, and reconstructed 3D environment data including their 3D semantically segmented counterparts.

\section*{Acknowledgements}

The work has been funded by the Deutsche Forschungsgemeinschaft (DFG, German Research Foundation) FOR 2535 Anticipating Human Behavior.

\bibliographystyle{abbrv}
\bibliography{references}

\end{document}